\pdfoutput=1
\PassOptionsToPackage{table}{xcolor}
\documentclass[fleqn,10pt]{wlscirep}
\usepackage[utf8]{inputenc}
\usepackage{array}
\usepackage{pdflscape}
\usepackage{longtable}
\usepackage{threeparttable}
\usepackage{graphicx}
\definecolor{heatmap1}{RGB}{8,48,107}    % Dark Blue (High value)
\definecolor{heatmap2}{RGB}{33,113,181}  % Medium Blue
\definecolor{heatmap3}{RGB}{66,146,198}  % Light Blue
\definecolor{heatmap4}{RGB}{107,174,214} % Lighter Blue
\definecolor{heatmap5}{RGB}{198,219,239} % Very Light Blue
\definecolor{heatmap6}{RGB}{237,248,251} % Almost White

\usepackage[T1]{fontenc}
\usepackage{makecell}
\title{Exploration of Foundation Model-Based Robots in Patient and Elderly Care}

\author[1]{Zhiwen Qiu, M.S.}
\author[2*]{Wei Liu, Ph.D.}
\author[3*]{Yuexing Hao, Ph.D.}
\affil[1]{Cornell University, Ithaca, 14850, USA}
\affil[2]{Department of Radiation Oncology, Mayo Clinic, Phoenix, AZ, 85054, USA}
\affil[3]{Massachusetts Institute of Technology, Cambridge, MA, 02139, USA}

\affil[*]{Corresponding Author: yuexing@mit.edu, liu.wei@mayo.edu}

\begin{abstract}
Demand for older-adult and patient care is growing rapidly as populations age worldwide. Foundation models are increasingly being integrated into robots and interactive agents, with the promise of more flexible communication and personalized assistance. However, care settings require reliable and workflow-compatible systems with accountable human oversight, and it remains unclear whether current embodied systems can translate technical advances into clinical impact. This Perspective synthesizes foundation model-based care robots across three areas: design features, user experience, and evidence for care-related outcomes. Current systems most commonly use foundation models as conversational and reasoning layers within voice-centered socially assistive embodiments, while multimodal grounding and physical autonomy remain limited. Empirical evaluations report positive usability and engagement benefits, but reliability failures persist across the interaction pipeline such as hallucinations and conversational breakdowns. Evidence for care impact remains concentrated in proximal outcomes such as cognitive engagement and participation, with limited evidence for validated clinical or care-related changes. We argue that future research should transition toward care-specific evaluation standards, accountable autonomy, and integration into care workflows to support more responsive and responsible care technologies.
\end{abstract}


\begin{document}

\flushbottom
\maketitle
% \thispagestyle{empty}

\section{Introduction}
The global aging population is posing increasing demands on care systems, while clinical and healthcare professionals are facing growing workload constraints \cite{amuthavalli2022decade, grinin2023global, gutterman2023caregiving}. Robots that support routine communication, guidance, or physical assistance without explicit programming for daily scenarios can meaningfully extend capacities for caregivers. The rise of foundation models, especially large language models (LLMs) and vision-language models (VLMs), has therefore sparked interests in their applications to robotics for older-adult and patient care \cite{ahn2022can, zitkovich2023rt, hao2024advancing}. Compared to traditional rule-based dialogues and pre-defined interaction scripts, foundation models offer a practical way to enable more flexible communication and customized support in complex real-world care contexts \cite{xiao2025robot, hao2025personalizing}. In principle, foundation models trained on large-scale multimodal data such as text, images, and sensor streams hold the promise for robots to understand natural language instructions, adapt to unfamiliar situations, and interact more intuitively with people who need assistance \cite{achiam2023gpt, zhang2024vision}.

However, deploying such systems in care settings raises practical questions about reliability, safety, workflow compatibility, and whether the complexity they introduce actually resolves the issues caregivers and patients face \cite{roustan2025clinicians}. Although foundation models bring new capabilities that can potentially generalize across interactive tasks, learn from demonstrations, and handle human environments without explicit feature engineering, they also introduce new challenges such as hallucinations, outdated information, and disruptions in multi-turn conversations \cite{irfan2025between, huang2025survey}. Despite rapid progress, much of the research focuses on technical capabilities and feasibility demonstrations, with less evidence that foundation model-based robots can deliver reliable clinical assistance in real-world care settings.

To address this gap, this Perspective examines how foundation models are currently integrated into embodied and semi-embodied care systems, how users experience these systems, and what kinds of health or care impact have actually been measured. We organize around three questions: (i) how foundation models change the design of care robots; (ii) what supports or limits user acceptance, engagement, and trust; and (iii) what evidence is available for care-relevant outcomes. We then discuss the main limitations of existing results and conclude by outlining future directions for evaluating these systems with stronger frameworks, accountable autonomy, and integration into real-world care workflows.

\begin{figure}[t]
    \centering
    \includegraphics[width=\linewidth]{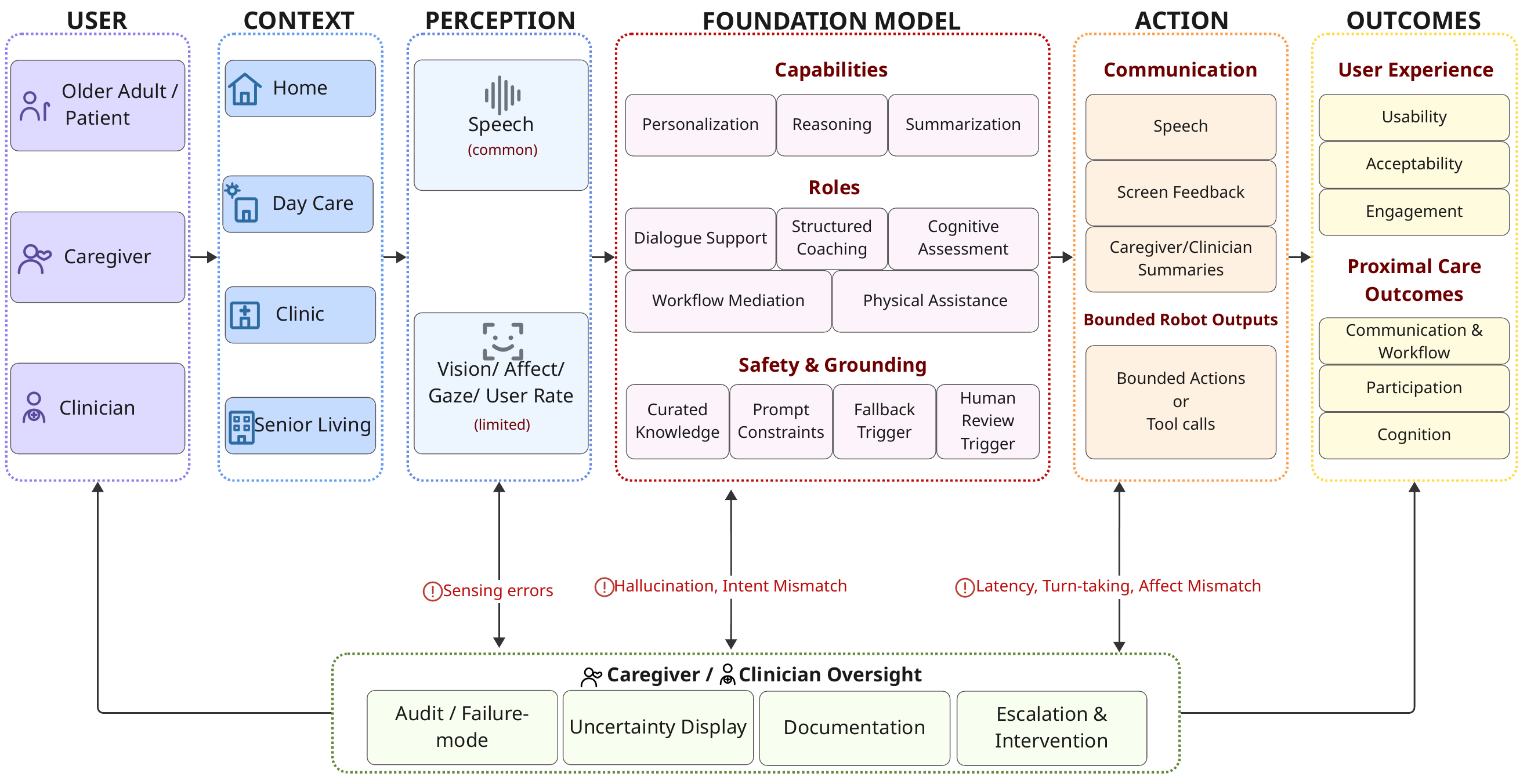}
    \caption{Interaction pipeline of foundation model-based care robots in patient and elderly care. Current systems commonly use foundation models as a conversational and reasoning layer between user input and robot output. Speech-based interaction is relatively mature, whereas multimodal perception, physical action, and standardized caregiver or clinician oversight remain less developed.}
    \label{fig:pipeline}
\end{figure}

\section{Features of Foundation Model-Based Care Robots}
\subsection{Roles of Foundation Models}
Foundation models appear in five major roles shown in ~Table \ref{tab:roles}. The most common role is open-ended dialogues where they generate personalized conversational content to support social connection, reminiscence, emotional expression, and everyday engagement \cite{khoo2023spill, irfan2024recommendations, irfan2025between, pinto2025designing}. In these studies, foundation model often functions as the conversational agent that enables more flexible turn-taking, personalized responses, and longer open-domain interactions compared to conventional scripted systems. The second role focuses on structured coaching and therapy, in which prompts and dialogue management deliver specific interventions such as positive psychology coaching \cite{spitale2025vita}, motivational reflection \cite{browne2024reflective}, range-of-motion guidance \cite{miyake2023feasibility}, or storytelling-based cognitive stimulation \cite{blanco2024storytelling}. The third role is assessment and screening. Examples include multi-LLM conversational pipelines for dementia-related monitoring \cite{numao2025interactive} and Blossom-based robots that administer cognitive tasks adapted from clinical instruments \cite{lima2025promoting}. The fourth role is workflow mediation and care coordination, where LLMs help translate user needs into structured actions or summaries, such as symptom reports and follow-up information for caregiver or clinician review \cite{yang2024talk2care}. A few studies also report robotic tool-use scenarios, where LLMs interpret high-level user commands and route them to bounded robot functions \cite{kang2024nadine, padmanabha2024voicepilot, kim2024framework}.

\subsection{Embodiment and Autonomy Levels}
Embodiment choices are highly concentrated. Most systems use socially assistive humanoids or tabletop robots, such as Pepper\cite{pandey2018mass}, NAO\cite{shamsuddin2011humanoid}, and Nadine\cite{ramanathan2019nadine}, where they primarily support social presence and engagement \cite{irfan2024recommendations, irfan2025between, blavette2025acceptability, lima2025promoting, blanco2024storytelling}. In these systems, foundation models are usually implemented as a conversational module over existing robotic platforms to enable more flexible dialogues, personalization, or task explanation while keeping the traditional action and control modules largely unmodified. This creates systems with relatively high conversational autonomy but limited grounded or physical autonomy. A smaller group of studies examines mobile companion robots that extend interaction into home-like or care environments \cite{logeshwar2025smart,blanco2025ebo,narraguide2025}. In these systems, mobility mainly supports navigation, social presence, and situated interactions and the foundation model is typically used for storytelling or conversational support. Physically assistive systems remain comparatively underexplored, with examples including assistive feeding, rehabilitation exercise guidance, and task-oriented robotic control \cite{padmanabha2024voicepilot,miyake2023feasibility,kim2024framework}. These systems generally use LLMs to interpret high-level user commands, generate verbal guidance, or support preference-sensitive interactions. Mobile and physically assistive embodiments broaden where and how care interactions can occur, but they have not yet produced continuous closed-loop autonomy in which robots adaptively perceive the environment, infer user state, and execute new physical behaviors in real time.

\subsection{Interaction Modality}
Interaction modality is strongly voice-centered. Most systems use speech as the primary input channel that combines automatic speech recognition, LLM-based response generation, and text-to-speech output to support real-time conversation. Vice is also paired with lightweight visual or physical feedback such as tablet screens, robot displays, confirmation prompts, or simple LED status indicators to help users understand whether the system is listening, processing, or awaiting confirmation \cite{blavette2025acceptability, vinay2025grace, favela2023conversational}. These hybrid interfaces are especially important in care contexts, where interaction failures may be caused not only by model inference errors, but also by hearing difficulty, speech recognition errors, cognitive load, or uncertainty about the robot's current state. 

Experimental comparisons across interaction modalities remain less studied. One example is Lima and colleagues'\cite{lima2025cultural} study which compared a smart speaker, a tablet-based virtual face, and an embodied robot in dementia-care interactions. They found high verbal engagement across all conditions (above 94\%), with higher cognitive difficulty for the embodied robot condition because participants needed to use an invocation name. The study suggested that additional interaction requirements such as remembering an invocation name, interpreting robot status, or coordinating speech with robot behavior, may increase cognitive burden even when the system appears more socially engaging. Some research also incorporate multimodal cues to facilitate interactions. Recent systems have utilized face tracking and emotion mirroring \cite{vantklooster2026gpt}, sentiment-aware dialogue logging \cite{huseynzade2025robots}, gaze-based agent selection in multi-agent reminiscence interaction \cite{sun2025chorus}, and webcam input to GPT-4o for co-creative drawing activities \cite{bossema2025llm}. These examples illustrate that foundation model-based care robots are beginning to integrate visual, affective, and behavioral contexts, but these cues are still mainly used to enrich prompts, personalize responses, or manage the interaction flow.

\begin{table}[t]
    \caption{Features of foundation models in care robots.}
    \label{tab:roles}
    \centering
    \small
    \setlength{\tabcolsep}{4pt}
    \renewcommand{\arraystretch}{1.2}
    \begin{tabular}{p{0.14\columnwidth} p{0.32\columnwidth} p{0.17\columnwidth} p{0.20\columnwidth} p{0.09\columnwidth}}
      \toprule
      \textbf{Role} & \textbf{Function} & \textbf{Embodiment} & \textbf{Autonomy Pattern} & \textbf{Examples} \\
      \midrule
      Conversation support &
      Generate flexible conversational content for social connection, reminiscence, and emotional expression &
      Socially assistive robot, tabletop robot, or humanoid robot &
      Conversational generation only &
      \cite{khoo2023spill, irfan2024recommendations, irfan2025between, pinto2025designing, huseynzade2025robots} \\
      \midrule
      Structured coaching &
      Deliver structured interventions such as well-being coaching, motivational reflection, storytelling, or cognitive stimulation &
      Socially assistive robot or mobile companion robot &
      Protocol-guided interaction &
      \cite{spitale2025vita, browne2024reflective, blanco2024storytelling, rincon2025personalized} \\
      \midrule
      Cognitive assessment &
      Administer cognitive tasks or estimate screening-related scores from interaction transcripts &
      Socially assistive robot or robot-based voice interface &
      Task-specific administration &
      \cite{numao2025interactive, lima2025promoting} \\
      \midrule
      Workflow mediation &
      Convert user input into clinician-facing summaries, constrained patient communication, or care coordination outputs &
      Robot-based voice interface, humanoid robot, or mobile robot &
      Human-reviewed output &
      \cite{yang2024talk2care, vantklooster2026gpt, kang2024nadine} \\
      \midrule
      Physical assistance &
      Interpret user preferences or high-level commands for rehabilitation, feeding, or task-oriented robot control &
      Physically assistive robot, humanoid robot, or manipulator &
      Constrained physical action &
      \cite{miyake2023feasibility, padmanabha2024voicepilot, kim2024framework} \\
      \bottomrule
    \end{tabular}
\end{table}

\subsection{Safety Mechanisms}
Safety mechanisms remain relatively modest in recent research. A common approach is prompt-level constraints incorporated with explicit non-medical-advice framing or restrictions on what the robot is allowed to recommend \cite{yang2024talk2care, padmanabha2024voicepilot}. Some systems add retrieval or curated knowledge sources to reduce unsupported outputs. For example, van 't Klooster and colleagues \cite{vantklooster2026gpt} whitelisted hospital information sources and used them to constrain GPT-generated responses for osteoarthritis communication. Other systems rely on intent classification and fallback behaviors, where uncertain user inputs are routed to safer responses instead of open-ended generation \cite{browne2024reflective}. Confirmation mechanisms are also present in workflow-mediating systems where user-provided symptom information is summarized instead of directly converted into medical advice \cite{yang2024talk2care}.

Structured safety designs are beginning to appear through modular or multi-agent architectures. For instance, the CORTEX architecture \cite{blanco2024storytelling} used in EBO storytelling decomposes interaction across dialogue, prompt generation, and emotion-recognition agents, while Numao and colleagues \cite{numao2025interactive} use four LangChain-based \cite{topsakal2023creating} LLM agents to separate free conversation from task-oriented cognitive assessment. These designs suggest the transition from single-model prompting toward more controlled interaction pipelines. However, end-to-end safety reporting remains limited as few studies systematically report failure modes or how human caregivers and clinicians should intervene when the system is uncertain.

\section{User Experience and Acceptance of Care Robots}
Empirical evidence for care robots centers on experiential evaluations that examine whether these systems are usable, acceptable, and engaging to older adults, patients, or healthcare professionals \cite{pinto2025designing, browne2024reflective}. These evaluations span home-like and senior-living environments \cite{numao2025interactive, lima2025promoting}, clinical and hospital contexts \cite{blavette2025acceptability, vantklooster2026gpt}, and controlled lab or simulation settings \cite{spitale2025vita, miyake2023feasibility}. Across contexts, foundation models improve user experience by enabling more conversational and adaptive interaction, but higher acceptance still depends on reliability, perceived safety, interaction burden, and satisfaction of user needs.

\subsection{Usability and Acceptability}
The clearest evidence for user-experience gains comes from research that compares care robots before and after foundation-model integration or iterative system refinement. Foundation models appear to improve acceptance mainly by making robot responses more coherent, contextual-aware, and responsive to user intents. Blavette and colleagues' \cite{blavette2025acceptability, blavette2025integrating} iterative socially assistive robot (SAR) deployment in a geriatric day-care unit showed substantial improvements in experiential metrics, with Acceptability E-scale scores increasing from 15.4 to 22.5 (out of 30) and System Usability Scale (SUS) \cite{lewis2018system} scores increasing from 47.9 to 69.3 (out of 100). The same deployment also reported fewer comprehension failures and higher interaction success after integrating a Vicuna-13B module, suggesting that language-model integration can improve not only subjective ratings but also the practical flow of interaction. Similar positive results were also reported in clinical communication and cognitive-stimulation contexts, including positive UEQ ratings from both osteoarthritis patients and healthcare professionals using a GPT-based robot, and above-midpoint ratings for ease of use, perceived alliance, and enjoyment after brief interactions with healthy older adults \cite{vantklooster2026gpt, vinay2025grace}. These findings suggest that users often respond positively when foundation models make robots more understandable and useful for the specific interaction demands of a care context.

% Similar positive results were also reported in clinical settings. Van 't Klooster and colleagues \cite{vantklooster2026gpt} reported positive User Experience Questionnaire (UEQ) scores averaging 2.13 (out of 3) for both osteoarthritis patients and healthcare professionals when a GPT-based robot was used to support clinical communication. Vinay and colleagues' \cite{vinay2025grace} pilot study similarly reported above-midpoint ratings for ease of use, perceived alliance, and enjoyment after a brief interaction with healthy older adults. 

\subsection{Engagement and Personalization}
Empirical results in engagement is less standardized but demonstrates a broader range of interaction benefits. Several studies suggest that foundation model-based robots can sustain attention, verbal participation, and willingness to continue interaction, especially when activities such as conversation, reminiscence, storytelling, or cognitive stimulation are involved \cite{rincon2025personalized, vinay2025grace, blanco2024storytelling}. Storytelling and coaching systems can also make interactions more socially and cognitively engaging. Blanco and colleagues \cite{blanco2024storytelling} evaluated the EBO storytelling system with therapists, who rated it positively for dialogue naturalness, Mini-Mental State Examination (MMSE) adaptability, and cognitive-function-training potential. Older adults also report high enjoyment and willingness to replay the activity. Evidence from adaptive coaching further suggests that users prefer interactions that respond to their inputs and maintain a more personalized conversational flow \cite{spitale2025vita}. As a result, engagement depends less on embodiment alone but also on the abilities to sustain coherent and responsive conversations.

Personalization is used to make care-robot interactions more continuous, adaptive, and relevant to individual users. Pinto and colleagues \cite{pinto2025designing} described personalization through features such as memorizing prior interactions, adjusting turn-taking, using culturally appropriate voice styles, and selecting content based on user preferences. In dementia-care contexts, users and care professionals also emphasized the need for kind tone, appreciative language, and customizable expressions, while identifying challenges related to invocation names, accent recognition, and noisy environments \cite{lima2025cultural}. This implies that personalization should not only improve conversational fluency, but also reduce memory demands, avoid unnecessary interaction steps, and remain robust to diverse speech patterns and care settings. At the same time, personalization raises questions about user control. In Bossema and colleagues' co-creative drawing study \cite{bossema2025llm}, older adults preferred a curator role where they could evaluate and approve robot suggestions instead of fully delegating decisions to the robot. Personalization should therefore make interactions more adaptive while preserving user agency and allowing older adults or caregivers to guide the interaction.

\subsection{Reliability and Trust}
Reliability remains a major barrier to user trust in foundation model-based care robots. Irfan and colleagues \cite{irfan2025between} documented failure modes during open-domain GPT-3.5 dialogue with Swedish older adults, including frequent interruptions, slow or repetitive replies, incoherent responses, language errors, and outdated information. Similar barriers have also been noted in field deployments where many user inputs were routed to no-confidence fallback responses \cite{browne2024reflective}. These failures can occur across the interaction pipeline, including automatic speech recognition, LLM response generation, and text-to-speech output, and these breakdowns can reduce the robot's perceived persona and social consistency \cite{pinto2025designing}.

Reliability failures also come from affect, timing, and interaction management. For instance, Huseynzade and colleagues \cite{huseynzade2025robots} found that older adults generally remained engaged and described the android as a good listener, but occasional affective mismatches still disrupted rapport and highlighted the need for stronger timing and contextual control. Latency and speech breakdowns were also noted as interaction challenges in a generally well-received retirement-home deployment of socially assistive robots for cognitive-health tasks \cite{lima2025promoting}. For socially isolated or community-dwelling older adults, these issues are especially prominent because trust depends on natural turn-taking, emotionally appropriate responses, memory continuity, culturally aligned voice, and recovery from communication failures \cite{irfan2024recommendations, pinto2025designing}.

% Hallucination is therefore only one part of a broader reliability problem as trust depends on the full interaction pipeline rather than the foundation model alone.

\section{Clinical Outcomes}
Current evidence on care impact remains concentrated in proximal outcomes. Most studies evaluate whether foundation model-enabled care robots can support cognitive and behavioral care processes, such as improving communication, encouraging participation, or facilitating interaction, rather than demonstrating measurable changes in disease status, functional independence, or other downstream clinical outcomes. These outcomes are essential to indicate where near-term deployment is most feasible and clarify gaps between existing evidence and the level of validation required to support long-term clinical decision-making. The clearest evidence concerns care communication and workflow support. In these studies, the foundation model primarily functions as a language layer that transforms unstructured patient interaction into more structured, actionable information for patients, caregivers, or clinicians. For example, Yang and colleagues developed an LLM-based voice assistant that collects symptom reports from older adults and converts them into clinician-facing summaries to support asynchronous triage and follow-up \cite{yang2024talk2care}. Similarly, recent work also utilized GPT-augmented social robots for patient-facing clinical communication, including osteoarthritis education, with responses constrained by whitelisted hospital information sources \cite{vantklooster2026gpt}. Evidence from specialty-care patient education further illustrates the need for cautious deployment as LLM-generated content may be broadly accurate but the quality can vary substantially across clinical topics and therefore requires systematic auditing \cite{holmes2025radonc}.

Behavioral outcomes capture user engagement, participation, and task-related behavior during care activities. These outcomes assess whether users remain engaged, complete assigned activities, or demonstrate more appropriate task-relevant behaviors during longitudinal interactions with the robot. Repeated deployments suggest that adaptive interactions can maintain participation over time, as illustrated in a four-week robotic well-being coaching \cite{spitale2025vita} and five-week robot-administered cognitive-health sessions that produced enriched picture descriptions and fewer semantic-fluency repetitions \cite{lima2025promoting}. Task-specific evaluations show similar results that LLM-supported exercise guidance can improve touch-force appropriateness during range-of-motion practice, while an LLM-mediated speech interface can enable preference-sensitive control of an assistive feeding robot \cite{miyake2023feasibility,padmanabha2024voicepilot}.

Cognition-related evidence describes interaction-based indicators of cognitive engagement, memory recall, and screening-relevant behavior. They illustrate how these systems can structure cognitive activities, elicit memory-related responses, and derive preliminary monitoring signals from daily interactions. In dementia-care settings, robot-administered MMSE-anchored tasks also produced high attention and verbal engagement which suggests that cognitive-task participation can be sustained even among users with impairment \cite{lima2025cultural}. Other systems extend to monitoring and reminiscence. For example, multi-LLM dialogues have been used to estimate HDS-R scores from daily conversations \cite{numao2025interactive}, while ReminiBuddy showed that different memory cues elicited distinct forms of recall, with personal photos prompting autobiographical memories and nostalgic objects prompting era-related reflections \cite{sun2025chorus}. 

Current evidence suggests that the near-term value of foundation model-based care robots is in interaction-mediated care support than autonomous diagnosis or clinical treatment decision-making. These systems are beginning to show credible potential as communication mediators, adaptive scaffolds for care activities, and generators of cognition-relevant interactions. Future research is needed to investigate whether these proximal gains in information flow, participation, and cognitive engagement can be translated into durable benefits for patients, caregivers, and clinical teams.

\section{Limitations for Clinical Translation}
Although recent studies report encouraging outcomes, existing evidence remains insufficient for robust clinical translation. Two limitations are particularly notable. The first concerns evaluation design, including the selection of outcomes, the use of appropriate comparators, the duration of follow-up, and the extent to which measured effects are linked to clinical or caregiver decision-making. The second concerns evidence validity, including participant selection, deployment context, and whether the evaluated system represents a stable and reproducible technology rather than a prototype demonstration.

The first limitation is evaluation design. Current studies measure communication quality, usability, engagement, well-being, task completion, cognitive-task performance, and screening-adjacent estimates without a shared outcome framework. Findings such as HDS-R estimation \cite{numao2025interactive}, adaptive robotic coaching \cite{spitale2025vita}, asynchronous symptom triage \cite{yang2024talk2care}, and repeated cognitive tasks \cite{lima2025promoting} are meaningful within their respective contexts, but remain difficult to integrate into cumulative evidence. Evaluation duration is also limited. Many studies remain single-session feasibility evaluations, and even with longer deployments, including geriatric-care and cognitive-health studies, are short relative to the timescales of chronic care, dementia support, or caregiver burden \cite{blavette2025integrating,spitale2025vita,lima2025promoting}. Experiment design for comparators is similarly underdeveloped. Many studies lack non-LLM, human-administered, or standard-of-care baselines \cite{rincon2025personalized,kim2024framework,kang2024nadine}, which make it difficult to determine whether observed benefits arise from the foundation model, robot embodiment, novelty effects, or human facilitation. Moreover, proximal outcomes such as engagement, screening-adjacent estimates, or interaction quality are rarely linked to validated clinical endpoints or subsequent changes in care planning \cite{blanco2024storytelling}. Safety and accountability reporting remains an additional gap. Although many research describes prompt constraints, curated knowledge sources, confirmation or fallback behaviors, few report failure/near-miss rates, escalation frequency, or criteria for caregiver or clinician intervention \cite{vantklooster2026gpt,blavette2025acceptability}. Browne and colleagues’ \cite{browne2024reflective} reported that 47\% of logged intents fell into a no-confidence fallback is valuable precisely because this level of transparency remains uncommon. For foundation model-enabled robots to be evaluated as care technologies, failure-mode taxonomies, audit methods, uncertainty displays, and human-intervention rates should become standard evaluation outcomes.

The second limitation is evaluation validity. Current studies often evaluate systems with populations that approximate the intended users, such as non-disabled adults in assistive feeding, young adults in physical-care guidance, healthy workplace users in well-being coaching, or healthy older adults in cognitive-stimulation studies \cite{padmanabha2024voicepilot,miyake2023feasibility,spitale2025vita,vinay2025grace}. When target conditions are represented, they may be defined by screening instead of diagnostic confirmation \cite{lima2025promoting}, while some dementia-focused systems rely on synthetic personas or exclude people with dementia from evaluation altogether \cite{favela2023conversational,miyawaki2024development}. Samples are also frequently small, single site, and culturally narrowed, which limit claims about accessibility, and robustness for clinical translation \cite{lima2025cultural}. System implementation further constrains the validity of current evidence. Hallucinations, inappropriate responses, and interaction breakdowns continue to appear under realistic use conditions \cite{favela2023conversational,irfan2025between,vantklooster2026gpt,sun2025chorus,pinto2025designing}. A substantial number of current work also remains simulation-based, audio-only, architecture-focused, or supported by Wizard-of-Oz components for functions related to the claimed autonomy \cite{kim2024framework,khoo2023spill,blanco2025ebo}. Reproducibility is additionally complicated by reliance on proprietary models whose behavior may vary across versions, prompts, and deployment contexts \cite{miyake2023feasibility,browne2024reflective}. These constraints do not diminish the value of early prototypes, but define the boundary between feasibility and evidence for deployable care systems.

\section{Future Directions}
\subsection{Evaluation Standards}
Future evaluations need to transition from isolated feasibility demonstrations toward care-specific standards. Current studies often measure whether a robot is usable or engaging, but clinical translation requires outcome chains that connect interaction-level signals to downstream clinical outcomes. Communication quality should be linked to information completeness, caregiver burden, or visit quality. Sustained engagement should be connected to adherence or continued use, while cognition-relevant interaction signals should be tied to screening and monitoring. Metrics should therefore be organized around care goals such as communication support, cognitive stimulation, functional assistance, workflow integration, and safety.

Comparator design is equally important. Pre-registered comparisons against scripted dialogue, non-LLM systems, human-administered tasks, or robot-only embodiments would clarify what the foundation model itself contributes. Longer in-situ deployments are also needed to assess repeated use, caregiver involvement, and clinical handoff. Existing evaluation frameworks for clinical LLM agents offer useful starting points \cite{mehandru2024evaluating,holmes2025radonc}, but care robotics must extend them to account for embodiment, interaction failure, user vulnerability, and shared responsibility among patients and clinicians.

\subsection{Grounded and Accountable Autonomy}
Foundation models have expanded what care robots can say and summarize, but have not yet produced comparable advances in what robots can perceive, manipulate, or make decisions autonomously. Future systems need stronger links between conversational reasoning and embodied context. Speech, gaze, gesture, facial expression, environmental information, and user state should inform the robot's reasoning and behavior, rather than remaining auxiliary inputs used only to enrich prompts. Greater autonomy should also be paired with stronger accountability. Near-term care robots are likely to operate through bounded shared-autonomy mechanisms, where the robot acts within a constrained skill library but defers when confidence, grounding, or safety are insufficient. These systems should report failure-mode taxonomies, uncertainty displays, fallback rates, and human-intervention rates as evaluation outcomes. Reproducibility should also be treated as part of accountability by documenting model versions, prompt design, retrieval sources, and deployment contexts so that findings can accumulate despite rapid progress in foundation models.

\subsection{Care Workflow Integration}
Care robots are often evaluated as single-user systems rather than technologies embedded in care relationships among patients, caregivers, and clinicians. Future studies should therefore situate robots within workflows that specify what information is collected, who reviews it, when escalation occurs, and how responsibility is distributed across human and machines. This transition also requires more inclusive evaluation across cognitive impairment levels, languages, cultural contexts, sensory and motor abilities, and low-resource deployment settings. In populations with cognitive impairment, multi-user design is particularly important because caregivers may support consent, interpretation, follow-up, and sustained use. Regulatory questions should also be addressed earlier in the research process, including consent procedures, data governance, and liability across stakeholders and institutions. Treating these issues as design constraints rather than downstream barriers will help move foundation model-based care robots from promising interaction prototypes toward responsible care technologies.

\section{Code Availability}
No code was used for data processing or analysis.

\section{Acknowledgements}
This research was supported by the American Psychological Foundation K. Anders Ericsson Dissertation Research Grant, National Cancer Institute (NCI) R01CA280134, the Eric \& Wendy Schmidt Fund for AI Research and Innovation, the Fred C. and Katherine B. Anderson Foundation, and the Kemper Marley Foundation. The funders had no role in study design, data collection and analysis, decision to publish, or preparation of the manuscript.

\section{Author Contributions}
Z.Q. and Y.H. had full access to all of the data in the study and take responsibility for the integrity of the data and the accuracy of the data analysis.

The authors' contributions are as follows: Z.Q. and Y.H. conceived and designed the study, acquired and analyzed the data, and drafted the manuscript. Z.Q. and W.L. provided critical intellectual review and guidance. W.L. and Y.H. obtained funding and provided supervision. All authors contributed to administrative, technical, and material support, ensuring a comprehensive and collaborative research effort.

\section{Competing Interests}
The authors declare that they have no competing interests.

\bibliography{hrisurvey}

\end{document}